\documentclass[twoside,11pt]{article}

\usepackage{jmlr2e}

\usepackage{lastpage}
\usepackage{xspace}
\usepackage{courier}

\usepackage{multirow}
\usepackage{booktabs}

\def\eg{e.g.,~}

\newcommand{\appname}{\texttt{OnPrem.LLM}\xspace}

\usepackage{pifont}

\usepackage{balance}
\usepackage{graphicx}
\usepackage{caption}
\usepackage[list=true]{subcaption}

\usepackage{fancyvrb}

\usepackage{float}
\usepackage{listings}
\usepackage{color}
\usepackage{textcomp}
\definecolor{listinggray}{gray}{0.9}
\definecolor{lbcolor}{rgb}{0.9,0.9,0.9}
\renewcommand{\lstlistingname}{Example}
\usepackage{placeins}

\newcounter{Lcount}
\newcommand{\numsquishlist}{
  \begin{list}{\arabic{Lcount}. }
   { \usecounter{Lcount}
 \setlength{\itemsep}{0pt}      \setlength{\parsep}{3pt}
     \setlength{\topsep}{3pt}       \setlength{\partopsep}{0pt}
     \setlength{\leftmargin}{2.7em} \setlength{\labelwidth}{1em}
     \setlength{\labelsep}{0.5em} } }
\newcommand{\numsquishend}{\end{list}}

\newcommand{\squishlist}{
  \begin{list}{$\bullet$}
   { \setlength{\itemsep}{0pt}      \setlength{\parsep}{3pt}
     \setlength{\topsep}{3pt}       \setlength{\partopsep}{0pt}
     \setlength{\leftmargin}{1.5em} \setlength{\labelwidth}{1em}
     \setlength{\labelsep}{0.5em} } }
\newcommand{\squishend}{\end{list}}

\ShortHeadings{OnPrem.LLM: A Privacy-Conscious Document Intelligence Toolkit}{Maiya}
\firstpageno{1}

\hypersetup{ hidelinks }

\begin{document}

\title{OnPrem.LLM: A Privacy-Conscious\\Document Intelligence Toolkit}

\author{\name Arun S. Maiya \email amaiya@ida.org \\
       \addr Institute for Defense Analyses \\Alexandria, VA, USA}

\editor{~}

\maketitle

\begin{abstract}
We present \appname, a Python-based toolkit for applying large language models (LLMs) to sensitive, non-public data in offline or restricted environments. The system is designed for privacy-preserving use cases and provides prebuilt pipelines for document processing and storage, retrieval-augmented generation (RAG), information extraction, summarization, classification, and prompt/output processing with minimal configuration. \appname supports multiple LLM backends---including llama.cpp, Ollama, vLLM, and Hugging Face Transformers---with quantized model support, GPU acceleration, and seamless backend switching. Although designed for fully local execution, \appname also supports integration with a wide range of cloud LLM providers when permitted, enabling hybrid deployments that balance performance with data control. A no-code web interface extends accessibility to non-technical users.

\end{abstract}

\begin{keywords} generative AI, large language models, LLM, NLP, machine learning \end{keywords}

\section{Introduction}  \label{introduction}
Large language models (LLMs) such as GPT-4, LLaMA, Mistral, and Claude have significantly advanced the state of natural language processing, exhibiting strong performance across tasks including text generation, summarization, question answering, and code synthesis  \citep{anthropic2024claude3,jiang2023mistral7b,openai2024gpt4technicalreport,touvron2023llamaopenefficientfoundation}. While many of these models are accessible through cloud-based APIs, it is difficult to apply them to sensitive, non-public data in regulated or restricted environments.

Organizations in domains such as defense, healthcare, finance, and law often operate under strict data privacy and compliance requirements. These constraints frequently prohibit the use of general-purpose external services, particularly in environments with firewalls, air-gapped networks, or classified (or otherwise sensitive) workloads. 

Common approaches to LLM deployment under these constraints include: (1) running lightweight, local models with reduced performance; (2) self-hosting larger models at significant infrastructure cost; or (3) implementing complex, hybrid pipelines to leverage cloud APIs without violating data governance. Each path incurs trade-offs in accuracy, scalability, latency, and maintainability. Prior work on privacy-focused systems (\eg PrivateGPT, LocalGPT, GPT4All) typically lack comprehensive end-to-end pipelines for common tasks, do not support a diverse range of LLM backends,  offer minimal no-code options for non-technical users, and are often constrained by a narrow focus on specific retrieval strategies  \citep{anand2023gpt4allecosystemopensource,privategpt2023,localgpt2023}.

To address this gap, we introduce \appname, a modular, production-ready toolkit for applying LLMs to private document workloads in constrained environments. The system supports both fully local execution and secure integration with privacy-compliant cloud endpoints, giving organizations flexible control over data locality and model placement. It includes a suite of prebuilt pipelines for common document intelligence tasks such as advanced document processing (\eg table extraction, optical character recognition (OCR), markdown conversion), retrieval-augmented generation (RAG), information extraction, text classification, semantic search, and summarization.

\appname supports a range of LLM backends---including \textbf{llama.cpp} for efficient quantized inference, \textbf{Hugging Face Transformers} for broad model compatibility, \textbf{Ollama} for simplified local model orchestration, and \textbf{vLLM} for high-throughput, GPU-accelerated inference. In addition, the system provides optional connectors to privacy-compliant cloud providers (\eg AWS GovCloud, Azure Government).\footnote{\appname was originally named to reflect its exclusive focus on private, local LLMs. Support for cloud-based models was added later to enable hybrid local/cloud deployments and to support privacy-compliant cloud endpoints.} While the toolkit prioritizes privacy, publicly accessible cloud LLMs (e.g., OpenAI, Anthropic) can also easily be used for applications involving public or non-sensitive data (e.g., government policy documents, scientific publications), enabling hybrid deployments that balance performance with data control.  A unified API enables seamless backend switching, while a no-code web interface allows non-technical users to perform complex document analysis without programming. Key design principles include:

\squishlist
\item \textbf{Data control} -- Local processing by default; external access is opt-in and configurable
\item \textbf{Deployment flexibility} -- Operates on consumer-grade machines or GPU-enabled infrastructure
\item \textbf{Ease of integration} -- Python API, point-and-click web interface, and prebuilt workflows streamline setup and execution
\item \textbf{Real-world focus} -- Built-in pipelines solve practical document-centric tasks out of the box
\squishend

\appname is open-source, free to use under a permissive Apache license, and available on GitHub at: \url{https://github.com/amaiya/onprem}. The toolkit has been applied to a wide range of use cases in the public sector including horizon scanning of scientific and engineering research, analyses of government policy, qualitative survey analyses, and resume parsing for talent acquisition.

\section{Core Modules}  \label{modules}

\appname is organized into four primary modules that together provide a comprehensive framework for document intelligence:

\subsection{LLM Module}
The core engine for interfacing with large language models. It provides a unified API for working with various LLM backends including llama.cpp, Hugging Face Transformers, Ollama, vLLM, and a wide-range of cloud providers  \citep{anthropic2024claude3,llama_cpp_2023,ollama2025,kwon2023efficient,openai2024gpt4technicalreport,wolf2020huggingfacestransformersstateoftheartnatural}. This module abstracts the complexity of different model implementations through a consistent interface while handling critical operations such as model loading with inflight quantization support, easy accessibility to LLMs served through APIs, agentic-like RAG, and structured LLM outputs.

\subsection{Ingest Module}
A comprehensive document processing pipeline that transforms raw documents into retrievable knowledge. It supports multiple document formats with specialized loaders, automated OCR for image-based text, and extraction of tables from PDFs. The module offers three distinct vector storage approaches:

\numsquishlist
\item \textbf{Dense Store}: Implements semantic search using sentence transformer embeddings and ChromaDB for similarity-based retrieval using hierarchical navigable small-world (HNSW) indexes \citep{chroma2025}. Elasticsearch is also supported.
\item \textbf{Sparse Store}: Provides both on-the-fly semantic search and traditional keyword search through Whoosh\footnote{We use \texttt{whoosh-reloaded} (available at \url{https://github.com/Sygil-Dev/whoosh-reloaded}), a fork and continuation of the original Whoosh project. } (or Elasticsearch) with custom analyzers and custom fields.
\item \textbf{Dual Store}: Combines both approaches by maintaining parallel stores, enabling hybrid retrieval that leverages both semantic similarity and keyword (or field) matching.
\numsquishend

\subsection{Pipelines Module}
Pre-built workflows for common document intelligence tasks with specialized submodules:

\squishlist
\item \textbf{Extractor}: Applies prompts to document units (sentences/paragraphs/passages) to extract structured information with Pydantic model validation \citep{pydantic2025}.
\item \textbf{Summarizer}: Provides document summarization with multiple strategies including map-reduce for large documents and concept-focused summarization.\footnote{{\em Concept-focused summarization} is a technique available in \appname for summarizing documents with respect to a user-specified concept of interest.}
\item \textbf{Classifier}: Implements text classification through scikit-learn wrappers (\texttt{SKClassifier}), Hugging Face transformers (\texttt{HFClassifier}), and few-shot learning with limited examples (\texttt{FewShotClassifier})   \citep{pedregosa2011scikit,tunstall2022efficientfewshotlearningprompts,wolf2020huggingfacestransformersstateoftheartnatural}.
\item \textbf{Agent}: Builds LLM-powered agents to execute complex tasks using tools and methods.
\squishend

\subsection{App Module}
\begin{figure}[hbt!]
  \centering
  \includegraphics[width=\textwidth]{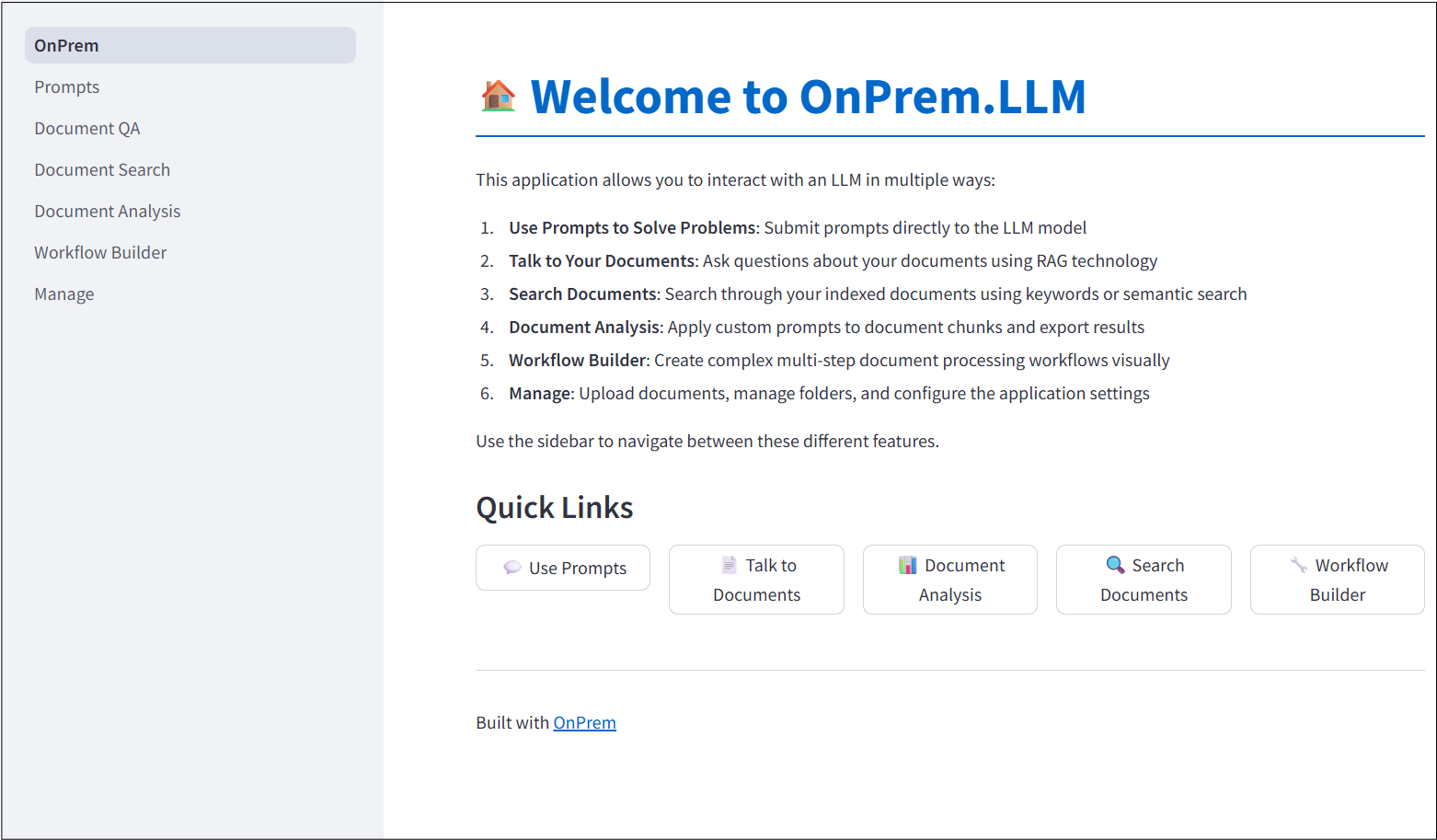}
  \caption{The \appname Web UI.}
  \label{fig:webui}
\end{figure}

As shown in Figure \ref{fig:webui}, a Streamlit-based web application makes the system accessible to non-technical users through six specialized interfaces \citep{streamlit2025}. The web interfaces offer easy, point-and-click access to: 1) interactive chat with conversation history; 2) document-based question answering with source attribution to mitigate hallucinations; 3) keyword and semantic search with filtering, pagination, and result highlighting; 4) custom prompt application to individual document passages with Excel export capabilities; 5) a visual workflow builder for crafting more complex data analysis pipelines; and 6) an administrative interface for document ingestion, folder management, and application configuration. For more information, refer to the web UI documentation.\footnote{\url{https://amaiya.github.io/onprem/webapp.html}}

\section{Usage Examples}  \label{examples}

The package documentation provides numerous practical examples and applications.\footnote{\url{https://amaiya.github.io/onprem/}} To illustrate the software's ease of use, \lstlistingname~\ref{lst:rag} presents a concise example of retrieval-augmented generation (RAG). We download and ingest the House Report on the 2024 National Defense Authorization Act (NDAA) to answer a related policy question.

\begin{scriptsize}
\begin{lstlisting}[
    language=Python,
    float=htbp,
    caption={A Basic RAG Pipeline in \appname},
    label={lst:rag}
]
from onprem import LLM, utils

# STEP 1: download the 2024 NDAA report
url = 'https://www.congress.gov/118/crpt/hrpt125/CRPT-118hrpt125.pdf'
utils.download(url,'/tmp/ndaa2024/report.pdf')
 
# STEP 2: load the default LLM using the default LLM backend
llm = LLM(n_gpu_layers=-1)

# STEP 3: ingest documents into default vector store
llm.ingest('/tmp/ndaa2024')

# STEP 4: ask questions
result = llm.ask('What is said about hypersonics?')
print(f"ANSWER:\n{result['answer']}")

ANSWER:
The context provided highlights the importance of expanding programs related to 
hypersonic technology. The House Committee on Armed Services has directed the 
Secretary of Defense to ... (answer truncated due to space constraints)
\end{lstlisting}
\end{scriptsize}

\section{Conclusion}

\appname addresses the critical need for privacy-preserving document intelligence in restricted environments. By combining local LLM inference, cloud-based LLM options, and a modular architecture with prebuilt pipelines, the toolkit enables organizations to implement advanced NLP workflows without compromising data governance. Its design allows for flexible control over the system footprint, making it suitable for a variety of application environments. As LLMs continue to improve in capability and efficiency, the demand for frameworks that support privacy-conscious, resource-aware use will only grow.

\FloatBarrier 
\begin{normalsize}
\vskip 0.2in
\bibliography{main}

\begin{thebibliography}{16}
\providecommand{\natexlab}[1]{#1}
\providecommand{\url}[1]{\texttt{#1}}
\expandafter\ifx\csname urlstyle\endcsname\relax
  \providecommand{\doi}[1]{doi: #1}\else
  \providecommand{\doi}{doi: \begingroup \urlstyle{rm}\Url}\fi

\bibitem[Anand et~al.(2023)Anand, Nussbaum, Treat, Miller, Guo, Schmidt,
  Community, Duderstadt, and Mulyar]{anand2023gpt4allecosystemopensource}
Yuvanesh Anand, Zach Nussbaum, Adam Treat, Aaron Miller, Richard Guo, Ben
  Schmidt, GPT4All Community, Brandon Duderstadt, and Andriy Mulyar.
\newblock Gpt4all: An ecosystem of open source compressed language models.
\newblock 2023.
\newblock URL \url{https://arxiv.org/abs/2311.04931}.

\bibitem[{Anthropic}(2024)]{anthropic2024claude3}
{Anthropic}.
\newblock The claude 3 model family: Opus, sonnet, haiku.
\newblock \url{https://www.anthropic.com/claude-3-model-card}, 2024.
\newblock Accessed: 2025-05-07.

\bibitem[Colvin et~al.(2025)Colvin, Jolibois, Ramezani, Badaracco, Dorsey,
  Montague, Matveenko, Trylesinski, Runkle, Hewitt, Hall, and
  Plot]{pydantic2025}
Samuel Colvin, Eric Jolibois, Hasan Ramezani, Adrian~Garcia Badaracco, Terrence
  Dorsey, David Montague, Serge Matveenko, Marcelo Trylesinski, Sydney Runkle,
  David Hewitt, Alex Hall, and Victorien Plot.
\newblock Pydantic.
\newblock \url{https://github.com/pydantic/pydantic}, April 2025.
\newblock URL \url{https://docs.pydantic.dev/latest/}.
\newblock MIT License. Accessed: 2025-05-07.

\bibitem[{Gerganov et al.}(2023)]{llama_cpp_2023}
{Gerganov et al.}
\newblock llama.cpp: Llm inference in c/c++.
\newblock \url{https://github.com/ggml-org/llama.cpp}, 2023.
\newblock Accessed: 2025-05-07.

\bibitem[Huber et~al.(2025)Huber, Troynikov, and Contributors]{chroma2025}
Jeff Huber, Anton Troynikov, and Chroma Contributors.
\newblock Chroma: The ai-native open-source embedding database.
\newblock \url{https://github.com/chroma-core/chroma}, 2025.
\newblock Version 1.0.8. Accessed: 2025-05-07.

\bibitem[{Iván Martínez}(2023)]{privategpt2023}
{Iván Martínez}.
\newblock Privategpt.
\newblock May 2023.
\newblock URL \url{https://github.com/zylon-ai/private-gpt}.
\newblock Software available at \url{https://www.zylon.ai/}.

\bibitem[Jiang et~al.(2023)Jiang, Sablayrolles, Mensch, Bamford, Chaplot,
  de~las Casas, Bressand, Lengyel, Lample, Saulnier, Lavaud, Lachaux, Stock,
  Scao, Lavril, Wang, Lacroix, and Sayed]{jiang2023mistral7b}
Albert~Q. Jiang, Alexandre Sablayrolles, Arthur Mensch, Chris Bamford,
  Devendra~Singh Chaplot, Diego de~las Casas, Florian Bressand, Gianna Lengyel,
  Guillaume Lample, Lucile Saulnier, Lélio~Renard Lavaud, Marie-Anne Lachaux,
  Pierre Stock, Teven~Le Scao, Thibaut Lavril, Thomas Wang, Timothée Lacroix,
  and William~El Sayed.
\newblock Mistral 7b.
\newblock 2023.
\newblock URL \url{https://arxiv.org/abs/2310.06825}.

\bibitem[Kwon et~al.(2023)Kwon, Li, Zhuang, Sheng, Zheng, Yu, Gonzalez, Zhang,
  and Stoica]{kwon2023efficient}
Woosuk Kwon, Zhuohan Li, Siyuan Zhuang, Ying Sheng, Lianmin Zheng, Cody~Hao Yu,
  Joseph~E. Gonzalez, Hao Zhang, and Ion Stoica.
\newblock Efficient memory management for large language model serving with
  pagedattention.
\newblock In \emph{Proceedings of the ACM SIGOPS 29th Symposium on Operating
  Systems Principles}, 2023.
\newblock URL \url{https://github.com/vllm-project/vllm}.

\bibitem[{Ollama Contributors}(2025)]{ollama2025}
{Ollama Contributors}.
\newblock Ollama.
\newblock \url{https://github.com/ollama/ollama}, 2025.
\newblock Accessed: 2025-05-07.

\bibitem[{OpenAI et al.}(2024)]{openai2024gpt4technicalreport}
{OpenAI et al.}
\newblock Gpt-4 technical report.
\newblock 2024.
\newblock URL \url{https://arxiv.org/abs/2303.08774}.

\bibitem[Pedregosa et~al.(2011)Pedregosa, Varoquaux, Gramfort, Michel, Thirion,
  Grisel, Blondel, Prettenhofer, Weiss, Dubourg, et~al.]{pedregosa2011scikit}
Fabian Pedregosa, Ga{\"e}l Varoquaux, Alexandre Gramfort, Vincent Michel,
  Bertrand Thirion, Olivier Grisel, Mathieu Blondel, Peter Prettenhofer, Ron
  Weiss, Vincent Dubourg, et~al.
\newblock Scikit-learn: Machine learning in python.
\newblock \emph{Journal of machine learning research}, 12\penalty0
  (Oct):\penalty0 2825--2830, 2011.

\bibitem[PromptEngineer(2023)]{localgpt2023}
PromptEngineer.
\newblock localgpt.
\newblock \url{https://github.com/PromtEngineer/localGPT}, 2023.
\newblock Accessed: 2025-05-07.

\bibitem[{Streamlit Contributors}(2025)]{streamlit2025}
{Streamlit Contributors}.
\newblock Streamlit: A faster way to build and share data apps.
\newblock \url{https://github.com/streamlit/streamlit}, 2025.
\newblock Accessed: 2025-05-07.

\bibitem[{Touvron et al.}(2023)]{touvron2023llamaopenefficientfoundation}
{Touvron et al.}
\newblock Llama: Open and efficient foundation language models.
\newblock 2023.
\newblock URL \url{https://arxiv.org/abs/2302.13971}.

\bibitem[Tunstall et~al.(2022)Tunstall, Reimers, Jo, Bates, Korat, Wasserblat,
  and Pereg]{tunstall2022efficientfewshotlearningprompts}
Lewis Tunstall, Nils Reimers, Unso Eun~Seo Jo, Luke Bates, Daniel Korat, Moshe
  Wasserblat, and Oren Pereg.
\newblock Efficient few-shot learning without prompts.
\newblock 2022.
\newblock URL \url{https://arxiv.org/abs/2209.11055}.

\bibitem[Wolf et~al.(2020)Wolf, Debut, Sanh, Chaumond, Delangue, Moi, Cistac,
  Rault, Louf, Funtowicz, Davison, Shleifer, von Platen, Ma, Jernite, Plu, Xu,
  Scao, Gugger, Drame, Lhoest, and
  Rush]{wolf2020huggingfacestransformersstateoftheartnatural}
Thomas Wolf, Lysandre Debut, Victor Sanh, Julien Chaumond, Clement Delangue,
  Anthony Moi, Pierric Cistac, Tim Rault, Rémi Louf, Morgan Funtowicz, Joe
  Davison, Sam Shleifer, Patrick von Platen, Clara Ma, Yacine Jernite, Julien
  Plu, Canwen Xu, Teven~Le Scao, Sylvain Gugger, Mariama Drame, Quentin Lhoest,
  and Alexander~M. Rush.
\newblock Huggingface's transformers: State-of-the-art natural language
  processing.
\newblock 2020.
\newblock URL \url{https://arxiv.org/abs/1910.03771}.

\end{thebibliography}
\end{normalsize}

\end{document}